\documentclass[letterpaper, 10 pt, conference]{ieeeconf}

\IEEEoverridecommandlockouts                              
\usepackage{amsmath,mathtools} 
\usepackage{amssymb}  
\usepackage{graphicx}
\usepackage[hidelinks]{hyperref}
\usepackage{siunitx}
\usepackage{subcaption}
\usepackage{booktabs}
\usepackage{xcolor}
\usepackage{textcomp}
\usepackage{gensymb} 
\usepackage{multirow} 
\usepackage{comment}
\usepackage{verbatim}
\usepackage[skip=0pt, font=footnotesize]{caption}
\usepackage{mathtools}
\usepackage{bm}
\usepackage[noadjust]{cite}
\usepackage{makecell}
\usepackage{tabularx}
\usepackage{pifont}
\usepackage{layouts}
\usepackage{physics}
\usepackage{algorithm}
\usepackage{algpseudocode}
\usepackage{layouts}
\usepackage{lipsum}

\DeclareRobustCommand{\rchi}{{\mathpalette\irchi\relax}}
\newcommand{\irchi}[2]{\raisebox{\depth}{$#1\chi$}} 
\newcommand{\txi}{{}_t\xi}
\newcommand\ringring[1]{%
  {
   \mathop{\kern0pt #1}\limits^{
     \vbox to-1.85ex{
       \kern-2ex 
       \hbox to 0pt{\hss\normalfont\kern.1em \r{}\kern-.45em \r{}\hss}%
       \vss 
     }
   }
  }
}

\DeclareRobustCommand{\bi}{\textbf{i}}
\DeclareRobustCommand{\bj}{\textbf{j}}
\DeclareRobustCommand{\bz}{\textbf{k}}
\DeclareRobustCommand{\bZ}{\bm{\zeta}}

\newcolumntype{M}[1]{>{\centering\arraybackslash}m{#1}}

\makeatletter
\def\endthebibliography{%
	\def\@noitemerr{\@latex@warning{Empty `thebibliography' environment}}%
	\endlist
}
\makeatother

\IEEEoverridecommandlockouts
\usepackage{tikz}
\usepackage{textcomp}
\usepackage{hyperref}
\usepackage{lipsum}

\newcommand\copyrighttext{%
	\footnotesize Published in IEEE Conference on Decision and Control (CDC), Cancun, Mexico, 2022.\newline
	 \textcopyright 2022 IEEE. Personal use of this material is permitted.
	Permission from IEEE must be obtained for all other uses, in any current or future media, including reprinting/republishing this material for advertising or promotional purposes, creating new collective works, for resale or redistribution to servers or lists, or reuse of any copyrighted component of this work in other works.}
\newcommand\copyrightnotice{%
	\begin{tikzpicture}[remember picture,overlay]
		\node[anchor=south,yshift=10pt] at (current page.south) {\fbox{\parbox{\dimexpr\textwidth-\fboxsep-\fboxrule\relax}{\copyrighttext}}};
	\end{tikzpicture}%
}

\title{\LARGE \bf
Spatial motion planning with Pythagorean Hodograph curves}

\author{Jon Arrizabalaga$^{1}$ and Markus Ryll$^{1,2}$
	\thanks{$^{1}$Autonomous Aerial Systems, School of Engineering and Design,  Technical University of Munich, Germany. E-mail: {\tt\small jon.arrizabalaga@tum.de} and {\tt\small markus.ryll@tum.de}}%
	\thanks{$^{2}$Munich Institute of Robotics and Machine Intelligence (MIRMI), Technical University of Munich}
}
\begin{document}
\maketitle
\copyrightnotice
\begin{abstract}
This paper presents a two-stage prediction-based control scheme for embedding the environment's geometric properties into a collision-free Pythagorean Hodograph spline, and subsequently finding the optimal path within the parameterized free space. The ingredients of this approach are twofold: First, we present a novel spatial path parameterization applicable to any arbitrary curve without prior assumptions in its adapted frame. Second, we identify the appropriateness of Pythagorean Hodograph curves for a compact and continuous definition of the path-parametric functions required by the presented spatial model. This dual-stage formulation results in a motion planning approach, where the geometric properties of the environment arise as states of the prediction model. Thus, the presented method is attractive for motion planning in dense environments. The efficacy of the approach is evaluated according to an illustrative example.
\end{abstract}
\vspace{1mm}
\textbf{Video}: \url{https://youtu.be/Sl6x8l7RJK8}

\section{INTRODUCTION}
\noindent Motion planning within cluttered and dynamic environments poses multiple challenges. Given a goal location and a performance criterion --duration, energy consumption or smoothness--, the underlying control scheme needs to drive the system, while remaining within a (possibly) variant free space. Thus, the optimal performance entails trading-off between the desired behavior, system constraints and spatial-awareness. To account for the latter, researchers have formulated path-parameterized control schemes allowing for a more precise embedding of the environment's geometric features. Such reformulations are based on a projection of the system dynamics from the Euclidean coordinate system to a moving frame attached to a path, located within the free space. The resulting system states --progress along the path and the orthogonal distance to it--, combined with the path's intrinsic properties --tangent, curvature and torsion-- arising from the parameterization, yield very appealing attributes not only for obstacle avoidance, but also for convergence to the desired performance criterion.

Advantages of converting time-dependent dynamics into spatial-dependent were initially discussed in \cite{pfeiffer1987concept}.  Applications of this spatial parameterization to planar vehicles demonstrated its ability to balance reference tracking and obstacle avoidance \cite{gao2012spatial, frasch2013auto}. Combining the spatial path-parameterization with advances in embedded-optimization allowed for real-time and near time-optimal Nonlinear Model Predictive Control (NMPC) applicable to miniature racing cars \cite{verschueren2014towards}. Subsequently, online obstacle avoidance was achieved in \cite{kloeser2020nmpc} by formulating a singularity-free parameterization of the system dynamics.

In the context of spatial dynamics, the expansion of the path-parameterization to all three dimensions has been simultaneously presented in \cite{van2016path, verschueren2016time, kumar2017path, spedicato2017minimum}. The first two works focus on robot manipulators and leverage the spatial parameterization in a path-following and a time-optimal motion planning NMPC schemes.  \cite{kumar2017path} exploits the decoupled tangential and transverse spatial states to formulate a controller for quadrotors capable of stabilizing the path following manifold, while \cite{spedicato2017minimum} aims for minimum-time and collision-free trajectory generation. Similarly, \cite{arrizabalaga2022towards} combines the time-optimality techniques from the planar methods with a complete spatial parameterization of the system dynamics to formulate a near time-optimal trajectory in real-time.

Given that the spatial path-parameterization conducted in all these works is based on the Frenet-Serret frame \cite{kuhnel2015differential}, the resultant equations of motion are not defined in inflection points, i.e., when the curvature vanishes, and thus, are only continuous for paths turning in one direction. Moreover, the undesired rotation of the Frenet-Serret frame with respect to its tangent component introduces a distortion in the representation of the environment \cite{wang2008computation}.
 
To account for these shortcomings, in this paper we derive a spatial path-parameterization applicable to any adapted frame, and thus, resulting in a generalization of the state of the art's equations of motion. Subsequently, we present a two-stage motion planning approach, where the first stage embeds the environment's geometry into a collision-free Pythagorean Hodograph spline, compliant with the aforementioned parameterization, while the second stage finds the (local) optimal path according to a performance criterion, the parameterized system dynamics and space constraints. In particular, we make the following contributions:
\begin{enumerate}
    \item We parameterize the three-dimensional Euclidean coordinates with respect to a path with an \emph{arbitrary adapted frame}. To the best of the authors' knowledge, this is the first spatial path-parameterization that is independent from the Frenet-Serret frame.
    \item  We identify the suitability of Pythagorean Hodograph curves \cite{farouki2008pythagorean} to efficiently and continuously define the parametric-speed, adapted frame components and angular velocity needed by the aforementioned parametrization. 
    \item We present a hierarchical motion planning algorithm in which the spatial features of the environment are first encoded into a Pythagorean Hodograph spline and then exploited in a prediction-based optimization. 
\end{enumerate}
\newpage
The remainder of this paper is structured as follows: Section~\ref{sec:spatial_path_parameterization} path parameterizes the three-dimensional Euclidean coordinates for an arbitrary adapted frame. Section~\ref{sec:pythagorean_hodograph_curves} introduces Pythagorean Hodograph curves and exhibits their applicability to the derived parameterization. Section~\ref{sec:solution_approach} presents the hierarchical motion planner and the respective two optimization problems. Experimental results are shown in Section~\ref{sec:experimental_results} before Section~\ref{sec:conclusion} presents the conclusions.

\hfill

\noindent\textit{Notation:} We will use $\dot{(\cdot)} = \dv{(\cdot)}{t}$ for time derivatives and $(\cdot)' = \dv{(\cdot)}{\xi}$ for differentiating over path-parameter $\xi$. For readability we will employ the abbreviation $ \txi=\xi(t)$.

\section{SPATIAL PATH-PARAMETERIZATION}\label{sec:spatial_path_parameterization}
\subsection{Preliminaries on space curves}
\noindent Let $\Gamma$ be a curve whose position and orientation are given by two rational and sufficiently continuous functions that depend on path-parameter $\xi$:
\begin{equation}
\Gamma = \{\bm{\gamma}(\xi) \in \mathbb{R}^3, \text{R}(\xi) \in \mathbb{R}^{3x3}\,|\,\xi \in[0,1] \}
\end{equation}
The rational orthonormal frame $\text{R}(\xi) = \left[\bm{e_1}(\xi),\bm{e_2}(\xi),\bm{e_3}(\xi)\right]$ defines the orientation along the curve and is assumed to be \emph{adapted}, i.e., the first frame vector coincides with the curve tangent $\bm{e_1}(\xi) = \frac{\bm{\gamma}'(\xi)}{||\bm{\gamma}'(\xi)||_2}$. The change of this frame with respect to the path-parameter is specified by
\begin{gather}\label{eq:R_ode}
    \text{R}'(\xi) = \text{R}(\xi)\overbrace{\begin{bmatrix}
	0 & -\rchi_3(\xi) &\rchi_2(\xi) \\ \rchi_3(\xi) &0 &-\rchi_1(\xi)\\ -\rchi_2(\xi) &\rchi_1(\xi) &0
	\end{bmatrix}}^{\text{C}(\xi)}
\end{gather}
where $\text{C}(\xi)$ is the Cartan connection matrix associated to the angular velocity vector $\bm{\omega}(\xi)$
\begin{gather*}
    \bm{\omega}(\xi) = \rchi_1(\xi)\bm{e_1}(\xi) + \rchi_2(\xi)\bm{e_2}(\xi) + \rchi_3(\xi)\bm{e_3}(\xi).
\end{gather*}
From \eqref{eq:R_ode} its components are given by
\begin{subequations} \label{eq:angvel_comp}
\begin{gather}
	\rchi_1(\xi) = \bm{e_2}'(\xi)\,\bm{e_3}(\xi)\,,\\
	\rchi_2(\xi) = \bm{e_3}'(\xi)\,\bm{e_1}(\xi)\,,\\
	\rchi_3(\xi) = \bm{e_1}'(\xi)\,\bm{e_2}(\xi)\,.
\end{gather}
\end{subequations}
\subsection{Derivation of equations of motion}\label{subsec:deriv_ode}
\begin{figure}
	\centering
	\includegraphics[width=\linewidth]{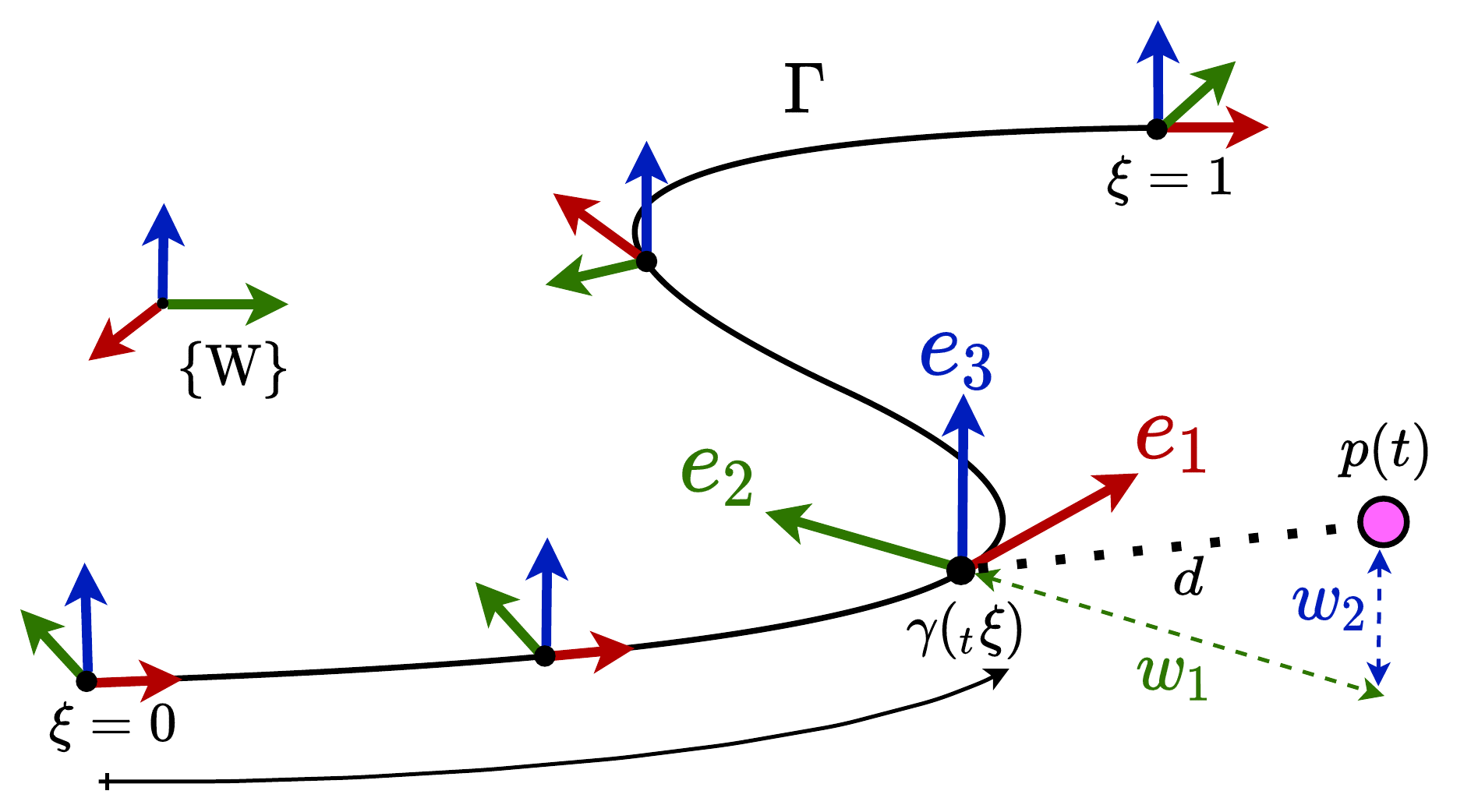}
	\caption{Spatial representation of a point-mass, represented by the pink dot, according to path-parameter $\xi$ and an adapted frame $\{\bm{e_1},\bm{e_2},\bm{e_3}\}$ of curve $\Gamma$. The distance $d(t)$ between the point-mass located in $p(t)$ and the closest point on the curve $\gamma({}_t\xi)$ is projected onto the transverse coordinates $w_1(t)$ and $w_2(t)$. For readability, the time dependencies on the distance and transverse coordinates have been omitted.}\label{fig:spatial_coordinates}
\end{figure}

\noindent Let $\bm{p}_\text{W}(t) \in \mathbb{R}^3$ be the location of a point-mass represented in world-frame's $(\cdot)_\text{W}$ Euclidean coordinates at time $t$. The distance with respect to the closest point on curve $\Gamma$ is given by $\bm{d}_\text{W}(t) = \bm{p}_\text{W}(t)-\bm{\gamma}(\txi)$. Translating this distance to the curve-frame $(\cdot)_\Gamma$ results in $\bm{d}_{\bm{\Gamma}}(t) = \text{R}(\txi)^\intercal \bm{d}_\text{W}(t)$, and therefore, its position in the world-frame can be denoted as
\begin{equation}\label{eq:pos_wf}
    \bm{p}_\text{W}(t) = \bm{\gamma}(\txi)+\text{R}(\txi)\bm{d}_{\bm{\Gamma}}(t)\;.
\end{equation}
Since we have assumed the frame $\text{R}(\txi)$ to be adapted, the first element of $\bm{d}_{\bm{\Gamma}}(t)$, i.e., the tangent component is zero, while the remaining two elements are the perpendicular projections, which we will refer to as \emph{transverse coordinates} $\bm{w}(t)=\left[w_1(t),w_2(t)\right]$. The projected distance can be observed in Fig.~\ref{fig:spatial_coordinates} and is expressed as
\begin{equation}\label{eq:dist_curveframe}
\bm{d}_{\bm{\Gamma}}(t) = \left[0,\bm{w}(t)\right] = \left[0, \bm{e_2}(\txi)\bm{d}_\text{W}(t), \bm{e_3}(\txi)\bm{d}_\text{W}(t)\right]^\intercal\,.
\end{equation}
Differentiating \eqref{eq:pos_wf} with respect to time results in
\begin{equation}\label{eq:vel_wf}
    \bm{v}_\text{W}(t) = \dot{\xi}(t)\left(\bm{\gamma}'(\txi)+\text{R}'(\txi)\bm{d}_{\bm{\Gamma}}(t)\right)+\text{R}(\txi)\dot{\bm{d}}_{\bm{\Gamma}}(t)\,.
\end{equation}
Denoting $\bm{i}_\text{W} = \left[1,0,0\right]^\intercal$ as the first component of the world-frame and setting the curve's parametric speed as $\sigma(\txi) = ||\bm{\gamma}'(\txi)||_2$ is equivalent to
\begin{gather}\label{eq:funcpos_der}
    \bm{\gamma}'(\txi)\equiv\sigma(\txi)\bm{e_1}(\txi)\,\equiv\,\text{R}(\txi) \bm{i}_\text{W}\sigma(\txi)\,.
\end{gather}
Introducing \eqref{eq:funcpos_der} in \eqref{eq:vel_wf} and multiplying it with $\text{R}^\intercal(\txi)$ leads to
\begin{multline*}
    0 = \dot{\xi}(t)\left(
    \sigma(\txi)\bm{i}_\text{W}+
    \text{R}(\txi)^\intercal \, \text{R}'(\txi)\, \bm{d}_{\bm{\Gamma}}(\txi)
    \right)\\ + \dot{\bm{d}}_{\bm{\Gamma}}(t) - \text{R}^\intercal(\txi)\bm{v}_{\text{W}}(t)\,.
\end{multline*}
Leveraging \eqref{eq:R_ode}, the latter equation results in the following simplification
\begin{equation*}
    0 = \dot{\xi}(t)\left(
    \sigma(\txi)\bm{i}_\text{W}+
    \text{C}(\txi)\bm{d}_{\bm{\Gamma}}(\txi)
    \right)
    +
    \dot{\bm{d}}_{\bm{\Gamma}}(t) - \text{R}^\intercal(\txi)\bm{v}_{\text{W}}(t)\; ,
\end{equation*}
which combined with the Cartan matrix and~\eqref{eq:dist_curveframe} yields
\begin{subequations}\label{eq:point_mass_eq}
\begin{gather}
    \dot{\xi}(t) = \frac{\bm{e_1}(\txi)^\intercal \bm{v}_\text{W}(t)}{\sigma(\txi) - \rchi_3(\txi) w_1(t) + \rchi_2(\txi) w_2(t)}\,,\\
    \dot{w}_1(t) = \bm{e_2}(\txi)^\intercal \bm{v}_\text{W}(t) + \dot{\xi}(t) \rchi_1(\txi) w_2(t)\,,\\
    \dot{w}_2(t) = \bm{e_3}(\txi)^\intercal \bm{v}_\text{W}(t) - \dot{\xi}(t) \rchi_1(\txi) w_1(t)\, .
\end{gather}
\end{subequations}
These equations describe the motion of the \emph{spatial coordinates} $\xi$, $\omega_1$, $\omega_2$ under a world-frame velocity $\bm{v}_\text{W}$ with respect to curve $\Gamma$ whose parametric speed, adapted frame and angular velocity are $\sigma(\xi),\,\text{R}(\xi)$ and $\bm{\omega}(\xi)$.

\subsection{Comparison to Frenet-Serret based models}
\noindent It can be easily verified that the spatial path parameterization in \cite{van2016path, verschueren2016time, kumar2017path, spedicato2017minimum} is a particular case of equation~\eqref{eq:point_mass_eq}, where the adapted frame matches the Frenet-Serret frame, and thus, the components of its angular velocity are $\left[\rchi_1,\rchi_2,\rchi_3\right] = \left[\tau,0,\kappa\right]$, $\tau$ and $\kappa$ being the curve's torsion and curvature. Furthermore, if the curve is assumed to be parameterized directly by its arc-length $s=\xi$, the parametric speed reduces to a unit magnitude $\sigma(\xi) = \dv{s}{\xi}=1$. 

\subsection{Choosing an adapted frame}
\noindent The analytical simplifications of the Frenet-Serret frame come at the expense of 1) \emph{discontinuities} when the curvature vanishes $\kappa=0$, causing abrupt flips in the second and third components $\bm{e_2},\bm{e_3}$ of the frame and 2) an \emph{unnecessary twist} with respect to the first component $\bm{e_1}$, which occurs because $\rchi_2=0$ rotates the frame such that its second component $\bm{e_2}$ points towards the center of the curvature \cite{farouki2008pythagorean}.

\begin{figure}[!t]
	\includegraphics[width=0.49\linewidth]{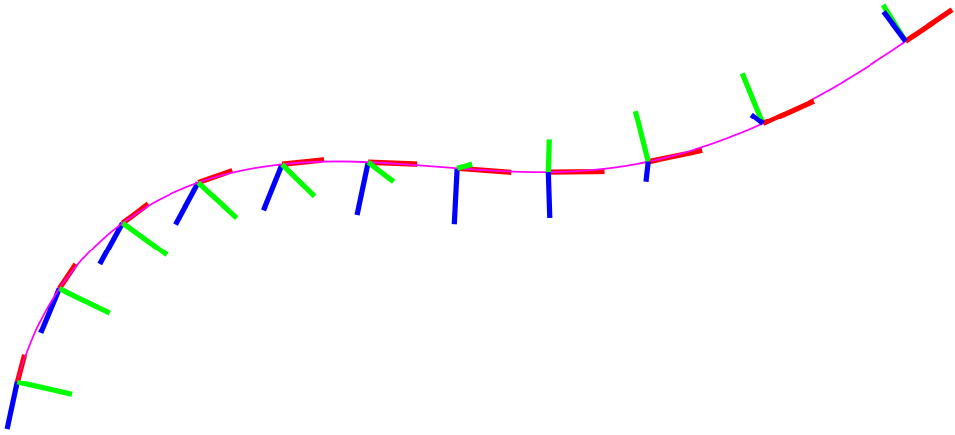}
	\includegraphics[width=0.49\linewidth]{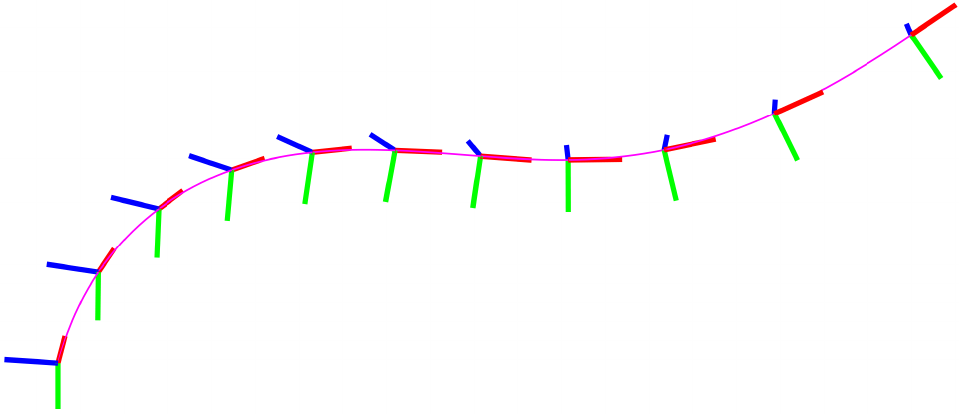}
	\vspace{3mm}
	\caption{A comparison between Frenet-Serret frame (left) and RMF (right) for a quintic Pythagorean Hodograph curve. Notice that the Frenet Serret frame shows an \emph{unnecessary rotation} with respect to the tangent component. This illustration is a recreation of the first example presented in \cite{farouki2012design}.}
	\label{fig:FS_vs_RMF}
\end{figure}

To ensure a continuous and smooth representation of the environment we are interested in finding a frame 1) that is \emph{defined} along the entire curve and 2) whose second and third components \emph{rotate the minimum possible amount} to ensure that the frame remains adapted. Such frames are denoted as \emph{Rotation Minimizing Frames} (RMF) \cite{bishop1975there} and are characterized by $\rchi_1 = 0$. An illustrative comparison between a Frenet-Serret frame and an RMF is depicted in Fig.~\ref{fig:FS_vs_RMF} .

\section{PYTHAGOREAN HODOGRAPH CURVES}\label{sec:pythagorean_hodograph_curves}
\noindent To leverage the derived spatial parameterization for prediction-based motion planning, we need a curve that 1) has an adapted frame which approximates an RMF, 2) has path-related functions $\sigma(\xi), \text{R}(\xi),\,\bm{\omega}({\xi})$ that can be expressed in closed-form by $C^2$ equations and 3) can be reconstructed using a small amount of parameters. The first requirement ensures a \emph{twist-free} and consistent encoding of the environment's properties, while the second and third allow for efficiently embedding~\eqref{eq:point_mass_eq} into a Nonlinear Program (NLP). 

A Pythagorean Hodograph (PH) curve is defined by the condition that the parametric speed $\sigma(\xi)$ is a polynomial of the path-parameter $\xi$ \cite{farouki2008pythagorean}, and thus, implies that $\sigma(\xi)$ meets conditions (2) and (3). In this section, we will see how PH curves also extend these requirements to the remaining two functions $\text{R}(\xi),\,\bm{\omega}({\xi})$, while simultaneously being able to compute RMF approximations.

\subsection{Preliminaries on Spatial PH curves}
\noindent Recalling that $\sigma(\xi) = ||\bm{\gamma}'(\xi)||_2$ and revisiting the aforementioned definition of PH curves results in
\begin{gather}\label{eq:ph_cond}
    \sigma^2(\xi) = {x'}^2(\xi)+ {y'}^2(\xi) + {z'}^2(\xi)\,,
\end{gather}
where $\sigma(\xi)$ is a polynomial. As proven in \cite{dietz1993algebraic}, every term in~\eqref{eq:ph_cond} can be expressed in terms of a quaternion polynomial
$\bm{Z}(\xi) = u(\xi) + v(\xi)\textbf{i} +g(\xi)\textbf{j} +h(\xi)\textbf{k}$, where $\{\textbf{i},\textbf{j},\textbf{k}\}$ refers to the $\mathbb{R}^{4}$ standard basis:
\begin{subequations}
\begin{gather}
    \sigma(\xi)  = u^2(\xi) + v^2(\xi) + g^2(\xi) + h^2(\xi)\,,\label{eq:sigma_sq}\\
    x'(\xi) = u^2(\xi) + v^2(\xi) - g^2(\xi) - h^2(\xi)\,,\\
    y'(\xi) = 2\left[u(\xi)h(\xi) + v(\xi)g(\xi)\right]\,,\\
    z'(\xi) = 2\left[v(\xi)h(\xi) - u(\xi)g(\xi)\right]\,
\end{gather}
\end{subequations}
Each component of the quaternion is a polynomial, and thus, can be expressed according to the Bernstein form as
\begin{gather} \label{eq:bernstein}
    \bm{Z}(\xi) = \sum_{i=0}^n \binom{n}{i}\bm{\zeta}_i\left(1-\xi\right)^{n-i}\xi^i\,,
\end{gather}
where $n$ refers to the degree of the polynomial and $\bm{\zeta}_i = \left[u_{i}, v_{i}, g_{i}, h_{i}\right] $ are the respective Bernstein coefficients. These relate to the control points of the quaternion polynomial $\bm{Z}(\xi)$ and may be rearranged into the following matrix:
\begin{gather}\label{eq:quat_controlpoints}
    \bm{\zeta} = \begin{bmatrix}u_{0}&v_{0}&g_{0}&h_{0}\\
    \multicolumn{4}{c}{\vdots}\\
    u_{n}&v_{n}&g_{n}&h_{n}\end{bmatrix}\,.
\end{gather}
From eqs. \eqref{eq:ph_cond}, \eqref{eq:sigma_sq} and \eqref{eq:bernstein} it can be concluded that a quaternion hodograph of degree $n$ corresponds to a curve of degree $2n+1$, while relying just on $4(n+1)$ Bernstein coefficients.

Another appealing feature of PH curves is their inheritance of a continuous adapted frame that is also solely dependent on its quaternion polynomial. This frame is named \emph{Euler Rodrigues Frame} (ERF) \cite{choi2002euler} and its respective components are described as
\begin{subequations}\label{eq:erf}
\begin{gather}
    \bm{e_1}(\xi) = \bm{Z}(\xi)\bi\bm{Z}^*(\xi)\,,\\
    \bm{e_2}(\xi) = \bm{Z}(\xi)\bj\bm{Z}^*(\xi)\,,\\
    \bm{e_2}(\xi) = \bm{Z}(\xi)\bz\bm{Z}^*(\xi)\,,
\end{gather}
\end{subequations}
with $(.)^*$ referring to the quaternion's conjugate. Combining this frame with~\eqref{eq:angvel_comp} leads to the following angular velocity components:
\begin{subequations}\label{eq:erf_angvel}
\begin{gather}
    \rchi_1 = \frac{2\left(uv'-u'v-gh'+g'h\right)}{u^2+v^2+g^2+h^2}\,,\label{eq:X1}\\
	\rchi_2 = \frac{2\left(ug'-u'g+vh'-v'h\right)}{u^2+v^2+g^2+h^2}\,,\\
	\rchi_3 = \frac{2\left(uh'-u'h-vg'+v'g\right)}{u^2+v^2+g^2+h^2}\,,
\end{gather}
\end{subequations}
where the dependency on path-parameter $\xi$ has been omitted for clarity. Putting together eqs. \eqref{eq:sigma_sq}, \eqref{eq:erf} and \eqref{eq:erf_angvel} it becomes apparent that PH curves enable us to express all path-related functions $\sigma(\xi)$, $\text{R}(\xi)$ and $\bm{\omega}(\xi)$ needed by the spatial parameterization in Section \ref{sec:spatial_path_parameterization} only depending on the Bernstein coefficients $\bm{\zeta}$ of the quaternion polynomial.

\subsection{PH curves with RMF}
\noindent As said before, an RMF is characterized for not rotating with respect to its tangent $\rchi_1=0$. In the case of PH curves, from~\eqref{eq:X1} results that the associated ERF is an RMF if the following condition holds true:
\begin{gather}\label{eq:rmf_ph_cond}
    u(\xi)v'(\xi)-u'(\xi)v(\xi)-g(\xi)h'(\xi)+g'(\xi)h(\xi) = 0
\end{gather}
Given its differential nature, a generic closed form solution is very difficult to obtain \cite{choi2002euler}. Nevertheless, in the upcoming section, this condition will play a crucial role when computing PH curves whose ERFs approximate RMFs.

\subsection{Spline of PH nonic curves}
\noindent To find a collision-free path, we are interested in concatenating multiple PH curves. When doing so, it has to be ensured that the path-related functions $\sigma(\xi)$, $\text{R}(\xi)$, $\omega(\xi)$ remain $C^2$. Out of all three, the frame's angular velocity is the one with a lowest degree of $n-1$, which is equivalent to a $C^3$ requirement in the quaternion polynomial. Thus, the minimum order for constructing PH splines under these conditions is $n=4$, resulting in curves of degree $9$, also known as \emph{nonics}.

Consequently, a spline with $m$ sections depends on $20m$ Bernstein coefficients. However, when applying the $C^{3}$ condition to the quaternion polynomials, only $4$ coefficients per section are independent, while the remaining $16$ are linearly related to the coefficients of the previous section: 
\begin{subequations}
    \begin{align}
    C^0&\xrightarrow{}&\,\bm{\zeta}_{0,k+1} &= \bm{\zeta}_{4,k},\\
    C^1&\xrightarrow{}&\,\bm{\zeta}_{1,k+1} &= -\bm{\zeta}_{3,k} + 2\bm{\zeta}_{4,k},\\
    C^2&\xrightarrow{}&\,\bm{\zeta}_{2,k+1} &= \bm{\zeta}_{2,k}-4\bm{\zeta}_{3,k}+4 \bm{\zeta}_{4,k},\\
    C^3&\xrightarrow{}&\,\bm{\zeta}_{3,k+1} &= -\bm{\zeta}_{1,k} + 6\bm{\zeta}_{2,k} - 12\bm{\zeta}_{3,k} + 8\bm{\zeta}_{4,k}\,.
\end{align}
\end{subequations}
for $k=1,...,m$ and $\bm{\zeta}_{i,k}$ referring to the control point $i$ --~row $i$ from the matrix in \eqref{eq:quat_controlpoints}~-- of the quaternion polynomial in section $k$. As a result of these equivalencies, a spline consisting of PH nonic curves depends on $20 + 4(m-1)$ Bernstein coefficients. 

Finally, the curve's control points $\bm{p}_{0,...,9|k}$ at the spline's section $k$ relate to the respective quaternion control points $\bm{\zeta}_{0,...,4|k}$ by the following expressions \cite{otto2021geometric}:
\begin{subequations}
\begin{align}
    \bm{p}_{1,k} &= \bm{p}_{0,k}+\frac{1}{9}\bZ_{0,k}\,\bi\,\bZ_{0,k}^*\,,\\
    \bm{p}_{2,k} &= \bm{p}_{1,k} + \frac{1}{18}\big(\bZ_{0,k}\,\bi\,\bZ_{1,k}^* + \bZ_{1,k}\,\bi\,\bZ_{0,k}^*\big) \,,\\
    \bm{p}_{3,k} &= \begin{multlined}[t]
                        \bm{p}_{2,k} + \frac{1}{126} \big(3\bZ_{0,k}\,\bi\,\bZ_{2,k}^* + 8\bZ_{1,k}\,\bi\,\bZ_{1,k}^* \\
                        +  3\bZ_{2,k}\,\bi\,\bZ_{0,k}^*\big)\,,
                   \end{multlined}\\
    \bm{p}_{4,k} &= \begin{multlined}[t]
                        \bm{p}_{3,k} + \frac{1}{126}\big(\bZ_{0,k}\,\bi\,\bZ_{3,k}^* + 6\bZ_{1,k}\,\bi\,\bZ_{2,k}^* \\
                        +  6\bZ_{2,k}\,\bi\,\bZ_{1,k}^* + \bZ_{3,k}\,\bi\,\bZ_{0,k}^*\big)\,,
                   \end{multlined}\\
    \bm{p}_{5,k} &= \!\begin{multlined}[t]
                        \bm{p}_{4,k} + \frac{1}{630}\big(\bZ_{0,k}\,\bi\,\bZ_{4,k}^* + 16\bZ_{1,k}\,\bi\,\bZ_{3,k}^* \\
                        + 36\bZ_{2,k}\,\bi\,\bZ_{2,k}^* +  16\bZ_{3,k}\,\bi\,\bZ_{1,k}^* \\+ \bZ_{4,k}\,\bi\,\bZ_{0,k}^*\big)\,,
                    \end{multlined}\\
    \bm{p}_{6,k} &= \begin{multlined}[t]
                        \bm{p}_{5,k} + \frac{1}{126}\big(\bZ_{1,k}\,\bi\,\bZ_{4,k}^* + 6\bZ_{2,k}\,\bi\,\bZ_{3,k}^* \\
                        +  6\bZ_{3,k}\,\bi\,\bZ_{2,k}^* + \bZ_{4,k}\,\bi\,\bZ_{1,k}^*\big)\,,
                   \end{multlined}\\
    \bm{p}_{7,k} &= \begin{multlined}[t]
                        \bm{p}_{6,k} + \frac{1}{126}\big(3\bZ_{2,k}\,\bi\,\bZ_{4,k}^* + 8\bZ_{3,k}\,\bi\,\bZ_{3,k}^* \\
                        +  3\bZ_{4,k}\,\bi\,\bZ_{2,k}^*\big)\,,
                   \end{multlined}\\
    \bm{p}_{8,k} &= \bm{p}_{7,k} + \frac{1}{18}\big(\bZ_{3,k}\,\bi\,\bZ_{4,k}^* + \bZ_{4,k}\,\bi\,\bZ_{3,k}^*\big) \,,\\
    \bm{p}_{9,k} &= \bm{p}_{8,k} + \frac{1}{9}\bZ_{4,k}\,\bi\,\bZ_{4,k}^*\,,
\end{align}
\end{subequations}
where $\bm{p}_{0,k}$ is a free integration constant that we set to the starting position of section $k$. With these control points the position function for a given section $k$ of the spline can be expressed in Bernstein's form as
\begin{gather}\label{eq:point_mass_pos}
\bm{\gamma}_k(\xi) = \sum_{i=0}^{9} \binom{9}{i}\bm{p}_{i,k}\left(1-\xi\right)^{9-i}\xi^i\,.
\end{gather}

\section{SPATIAL MOTION PLANNING} \label{sec:solution_approach}
\subsection{Stage 1: Computing collision-free PH nonic splines}
\noindent Other than being compact and continuous, PH splines also need to be collision-free. For this purpose, in a similar manner to \cite{liu2017planning}, we describe the non-convex free space $\mathcal{F}$ as the union of $m$ convex sets, each represented by a polyhedron, i.e., $\mathcal{F} = \cup_{k=1}^m\,\mathcal{P}_{k}$. Considering that a curve with a Bernstein Polynomial basis is contained in the convex hull of its control points, we can ensure the given PH spline section will be inside the respective polyhedron by requiring that all control points are enclosed within it. In section $k$, this translates to the following condition:
\begin{gather}\label{eq:ph_collisionfree}
    \textbf{A}_{k}\bm{p}_{i,k} \leq \textbf{b}_{k}\,\quad \text{with} \quad i = 0,...,9\,,
\end{gather}
where  $\bm{p}_{i,k}$ stands for the curve control points of the PH-spline's $k$-th section and ${\textbf{A}_{k},\textbf{b}_{k}}$ refer to the half-space representation of polyhedron $\mathcal{P}_k$. Under these constraints, the $20 + 4(m-1)$ free Bernstein coefficients can be used to frame the spline according to a desired criterion. Similarly to \cite{albrecht2020spatial}, acknowledging the lack of a closed-form solution to the differential condition in \eqref{eq:rmf_ph_cond}, we exploit the aforementioned degrees of freedom to find a PH nonic spline, whose ERF is as \emph{rotation minimizing as possible}. For a given section $k$, this is equivalent to minimizing the functional  
\begin{gather}\label{eq:ph_costfun}
f_{\text{PH}}(\bZ_{k}) = \int^1_0\rchi_{1,k}^2(\xi)\,d\xi\,.    
\end{gather} 
Combining~\eqref{eq:ph_costfun} with the curve's and quaternion polynomial's continuity conditions, as well as the collision-free constraints in \eqref{eq:ph_collisionfree}, results in the following optimal control problem (OCP):
\begin{subequations} \label{eq:ph_ocp}
\begin{alignat}{3}
&\min_{\bZ_{1},\cdots,\,\bZ_{m}} &&f_\text{PH}(\bZ_1,\cdots,\bZ_m)\\ 
&\quad\text{s.t.}&&\bm{p}_{0,1} = \bm{p}_\text{initial} \label{eq:ph_curve_init}\\
&&&\bm{p}_{9,m} = \bm{p}_\text{final} \label{eq:ph_curve_final}\\
&&&\bm{p}_{0,k+1} = \bm{p}_{9,k}\,,&\quad&k = 1,\cdots,m-1 \label{eq:ph_curve_cond}\\
&&&\bm{Z}_{k+1}(0) = \bm{Z}_{k}(1)\,, &\quad&k = 1,\cdots,m-1 \label{eq:ph_quat_cond_start}\\
&&&\bm{Z'}_{k+1}(0) = \bm{Z'}_{k}(1)\,, &\quad&k = 1,\cdots,m-1\\
&&&\bm{Z''}_{k+1}(0) = \bm{Z''}_{k}(1)\,, &\quad&k = 1,\cdots,m-1 \\
&&&\bm{Z'''}_{k+1}(0) = \bm{Z'''}_{k}(1)\,, &\quad&k = 1,\cdots,m-1 \label{eq:ph_quat_cond_end}\\
&&&\textbf{A}_{k}\bm{p}_{i,k} \leq \textbf{b}_{k}\,,&\quad&\substack{k = 1,\cdots,\,m\\i = 0,\cdots,9\,} \label{eq:ph_spatial_cond}
\end{alignat}
\end{subequations}
where \eqref{eq:ph_curve_init} and \eqref{eq:ph_curve_final} set the starting and ending point, \eqref{eq:ph_curve_cond} guarantees that the spline sections are attached to each other, from \eqref{eq:ph_quat_cond_start} to \eqref{eq:ph_quat_cond_end} ensure that the $\sigma(\xi)$, $\text{R}(\xi)$ and $\omega(\xi)$ functions are $C^2$, while \eqref{eq:ph_spatial_cond} enforces the spline to remain in the free space.

\subsection{Stage 2: Path-Parametric NMPC}
\noindent Once the OCP in \eqref{eq:ph_ocp} is solved, the parametric functions $\sigma(\xi)$, $\text{R}(\xi)$ and $\bm{\omega}(\xi)$ can be re-computed according to eqs. \eqref{eq:sigma_sq}, \eqref{eq:erf} and \eqref{eq:erf_angvel}. By embedding these functions into \eqref{eq:point_mass_eq}, we leverage the spatial reformulation in a prediction-based controller, where the environment's geometric properties are fully embedded into the system dynamics.

To do so, we take the standard NMPC approach \cite{camacho2013model}, where the input obtained from an optimization problem is applied in a receding horizon fashion. The respective OCP finds the local optimal control action over time horizon $T$, considering a nonlinear plant model $f(\bm{x},\bm{u})$, a nonlinear cost function $f_\text{MPC}(\bm{x},\bm{u})$, as well as nonlinear constraints $g(\bm{x},\bm{u})$ on states $\bm{x}$ and inputs $\bm{u}$. We denote the quaternion polynomial coefficients obtained from OCP \eqref{eq:ph_ocp} as $\bm{\mathcal{Z}}$. Choosing states $\bm{x} = \left[\xi,\bm{w}\right]$ and inputs $\bm{u} =\left[\bm{v}_\text{W}\right]$, the plant model is given by the equations of motion in~\eqref{eq:point_mass_eq}. As a consequence, the OCP that is solved in the motion planning's second stage is
\begin{subequations}\label{eq:mpc_ocp}
    \begin{flalign}
    &\min_{\bm{x}(\cdot),\bm{u}(\cdot)} \int_0^T f_\text{MPC}(\bm{x}(t),\bm{u}(t),\bm{\mathcal{Z}})\,dt&
    \end{flalign}
    \vspace{-5mm}
	\begin{alignat}{3}
	\text{s.t.}\quad& \bm{x}(0) = \bm{x}_{\text{initial}}\,,\\
	&\dot{\bm{x}} = f(\bm{x}(t),\bm{u}(t),\bm{\mathcal{Z}}), &\quad&t \in \left[0,T\right]\\
	&g\left(\bm{x}(t),\bm{u}(t),\bm{\mathcal{Z}}\right) \leq 0\,,    &\quad&t \in \left[0,T\right]\\
	&\textbf{A}_{\bm{x}}\,\bm{p}_{\text{W}}(\bm{x}(t),\bm{u}(t),\bm{\mathcal{Z}}) \leq \textbf{b}_{\bm{x}}\,,    &\quad&t \in \left[0,T\right] \label{eq:mpc_spatial_cond_cont}
	\end{alignat}
\end{subequations}
where $\textbf{A}_{\bm{x}}=\textbf{A}(\bm{x}(t))$ and $\textbf{b}_{\bm{x}}=\textbf{b}(\bm{x}(t))$ in constraint \eqref{eq:mpc_spatial_cond_cont} stand for the half-space matrixes of the polyhedron associated to the location of the point-mass $\bm{p}_{\text{W}}$, which can be computed from  \eqref{eq:pos_wf}. With this constraint, it is guaranteed that the motion planning takes place within the free space.
\subsection{Complete Approach}
\noindent The complete motion planning scheme, alongside an exemplary application within a generic control loop, is given in Algorithm \ref{alg:algorithm1}. After estimating the state of the system and decoupling the free space into multiple polyhedron, the motion planner solves both stages and finds the optimal path within a time horizon. In a receding horizon manner, only the first optimal input is applied. 

\begin{algorithm}[h]
\caption{Two-Stage Spatial Motion Planning}\label{alg:algorithm1}
\begin{algorithmic}[1]
\Function{Motion Planning}{$\bm{x}$, $\mathcal{P}_1,\cdots,\mathcal{P}_m$}
\State $\bm{p}_\text{init},\bm{p}_\text{final} \gets \text{FindStartAndEnd}(\mathcal{P}_1,\cdots,\mathcal{P}_m)$
\State $\bm{\mathcal{Z}} \gets \text{SolveStage1}(\bm{p}_\text{init},\bm{p}_\text{final},\mathcal{P}_1,\cdots,\mathcal{P}_m)$
\State $\bm{x}^*,\bm{u}^* \gets \text{SolveStage2}(\bm{x},\bm{\mathcal{Z}})$
\State\Return $\bm{x}^*,\bm{u}^*$
\EndFunction
\While{controller enabled \do}
\State $\bm{x} \gets \textsc{State Estimation}$
\State $\mathcal{P}_1,\cdots,\mathcal{P}_m \gets \textsc{Environment Mapping}$
\State $\bm{x^*},\bm{u^*}\gets\Call{Motion Planning}{\bm{x},\,
\mathcal{P}_1,\cdots,\mathcal{P}_m}$
\State $\textsc{Low Level Control}(\bm{u}^*_0)$
\EndWhile
\end{algorithmic}
\end{algorithm}

\begin{figure*}[t]
\centering
\includegraphics[width=\textwidth]{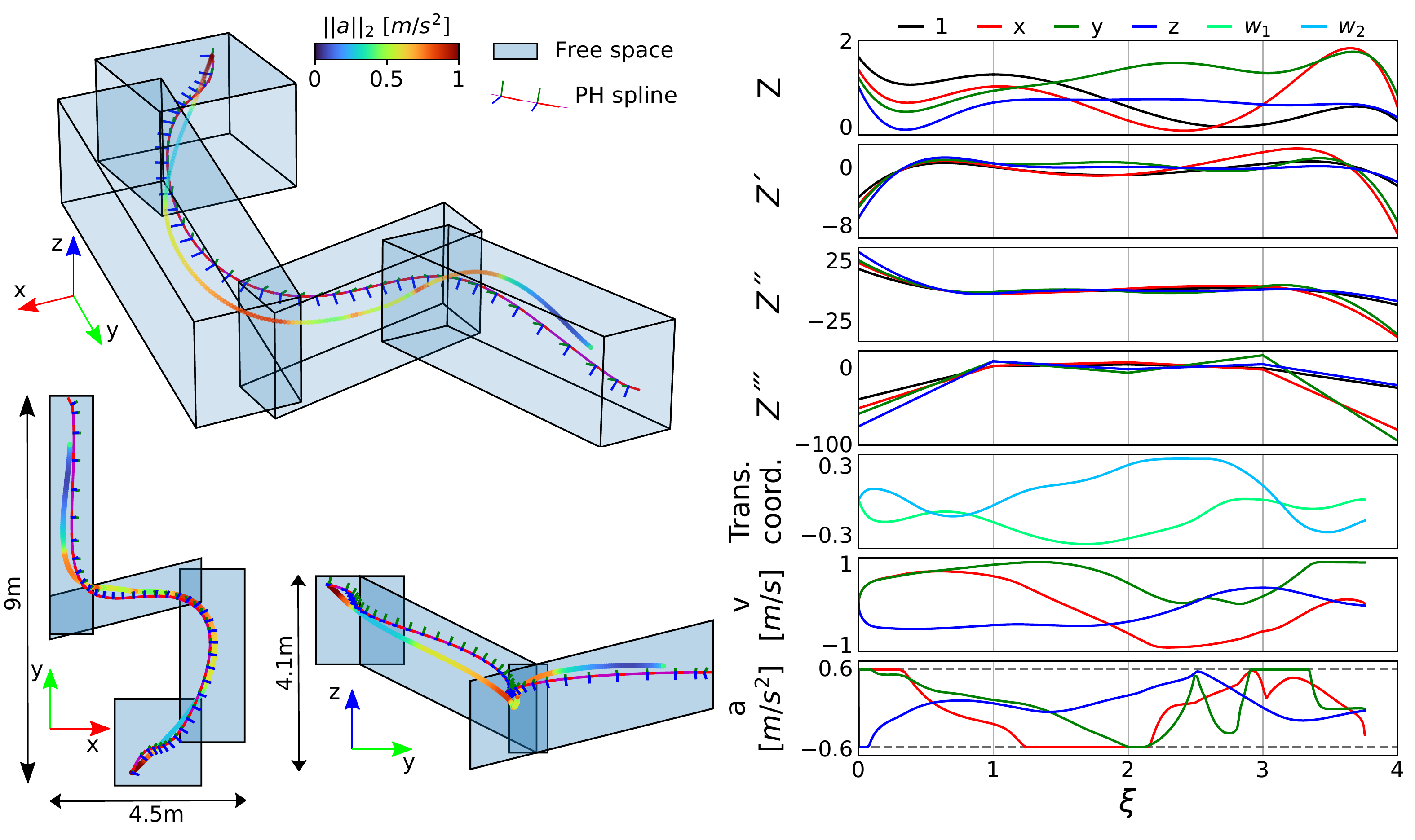}
\caption{Exemplary application of our two-stage spatial motion planning approach. \emph{Left column:} Starting from the top in clock-wise direction, isometric-view, side-view and top-view. The blue polyhedrons represent the free space, the PH spline is depicted according to its adapted frame and the planned path is given by the colored line. The color mapping relates to the norm of the acceleration. \emph{Right column:} The four divisions in the horizontal axis split the data according to the spline sections. The first four rows represent the quaternion polynomial and its derivatives, while the latter three rows show the system states (transverse coordinates and velocity), as well as the acceleration inputs.}\label{fig:results}
\end{figure*}

\section{TUTORIAL EXAMPLE}\label{sec:experimental_results}
\noindent To evaluate our approach, we generate an exemplary representation of the free space, consisting of four polyhedrons. Their sparsity along all three Euclidean axes, alongside their differences on size, allow for testing the capacity of our approach to deal with highly non-convex spaces. 

\subsection{Numerical implementation}
\noindent In the first stage, we formulate OCP \eqref{eq:ph_ocp} in CasADi \cite{andersson2019casadi} and solve it with IPOPT \cite{wachter2006implementation} and MA27 \cite{HSL} as the linear solver back-end. We approximate the integral in \eqref{eq:ph_costfun} with a 4th-order Runge-Kutta of step size $\Delta \xi = 0.1\,$. The decision variables are initialized by solving a least squares problem on the constraint-residuals with the Trust-Region-Reflective method \cite{conn2000trust}.

In the second stage, we approximate the OCP in \eqref{eq:mpc_ocp} by a Nonlinear Program (NLP) according to the multiple-shooting approach \cite{bock1984multiple}, in which time horizon $\text{T}$ is split into $\text{N}$ sections with constant decision variables:
\begin{subequations}\label{eq:mpc_NLP}
    \begin{flalign}
    &\min_{\substack{\bm{x}_{\,0},\cdots,\,\bm{x}_{\,N},\\\bm{u}_{\,0},\cdots,\,\bm{u}_{\,N}}}\sum_{k=0}^{N-1} f_{\text{MPC}.k}(\bm{x}_k,\bm{u}_k,\bm{\mathcal{Z}})&
    \end{flalign}
    \vspace{-5mm}
	\begin{alignat}{3}
	\text{s.t.}\quad& \bm{x}_{0} = \bm{x}_{\text{initial}}\,,\\
	&\bm{x}_{\,k+1} = F(\bm{x}_{k},\bm{u}_{k},\bm{\mathcal{Z}}, \Delta\,t), &\;&k = 0,\cdots,N-1\\
	&g\left(\bm{x}_{k},\bm{u}_{k},\bm{\mathcal{Z}}\right) \leq 0\,,    &\;&k = 0,\cdots,N-1 \label{eq:mpc_input_const}\\
	&\textbf{A}_{\bm{x}}\,\bm{p}_{\text{W}}(\bm{x}_{k},\bm{u}_{k},\bm{\mathcal{Z}}) \leq
	\textbf{b}_{\bm{x}}\,,    &\;&k = 0,\cdots,N-1 \label{eq:mpc_spatial_cond}\\
	&h\left(\bm{x}_{N},\bm{\mathcal{Z}}\right) \leq 0\,,    &\;&\label{eq:mpc_end_cond}
	\end{alignat}
\end{subequations}
with $\textbf{A}_{\bm{x}}=\textbf{A}(\bm{x}_k)$ and $\textbf{b}_{\bm{x}}=\textbf{b}(\bm{x}_k)$. The respective plant model $\text{F}$ is obtained from discretizing $f(\bm{x},\bm{u},\bm{\mathcal{Z}})$ with a fixed time-step $\Delta t$. As an illustrative showcase, the cost function is chosen to maximize for progress:
\begin{gather*}
f_{\text{MPC},k}(\bm{x}_k,\bm{u}_k,\bm{\mathcal{Z}}) = -\lambda\,L(\bm{x}_k,\bm{\mathcal{Z}}) + \left|\left|\bm{u}_{\,k}\right|\right|_{R}^2\, ,
\end{gather*}
where $L(\bm{x}_k,\bm{\mathcal{Z}})$ is the arc-length at state $\bm{x}_k$ and can be obtained by integrating 
\begin{gather*}
L(\bm{x}_k,\bm{\mathcal{Z}}) = \int^{\xi_k}_{0} \sigma(\xi)\, \text\,d\xi \,.
\end{gather*}
Given that the parametric speed $\sigma(\xi)$ is a polynomial (see~\eqref{eq:sigma_sq}), the integral above can be expressed in closed-form as function that is solely dependent on the quaternion coefficients $\bm{\mathcal{Z}}$ computed in the first stage. Moreover, to soften the control commands we extend the states by appending the velocity to the state vector $\bm{x} =\left[\xi,\bm{w},\bm{v}_{\text{W}}\right]$ and assigning the acceleration to the input $\bm{u} =\left[\bm{a}_{\text{W}}\right]$. As common, within \eqref{eq:mpc_input_const} we account for input constraints. Lastly, to ensure feasibility, we add an additional constraint \eqref{eq:mpc_end_cond} on the states of the last shooting node. 

We solve the NLP~\eqref{eq:mpc_NLP} with the Sequential Quadratic Programming (SQP) method in the optimal control framework ACADOS \cite{verschueren2018towards}. To address real-time applicability, we use its real-time iteration variant (SQP-RTI) \cite{diehl2005real} in conjunction with HPIPM \cite{frison2020hpipm}, which efficiently solves the underlying quadratic programs. The system dynamics are integrated by an explicit 4th-order Runge-Kutta method. 

We define a prediction horizon of \SI{2}{\second} with $40$ shooting nodes, equivalent to a sampling time of $50$ Hz. The weighting matrices are kept constant during all evaluations as $\lambda = 2$ and $R= 0.2I_3$, with $I_3$ referring to a $3 \times 3$ identity matrix. To resemble realistic motion dynamics, we limit the acceleration components to $\pm$\SI{0.58}{\meter/\second^2}, i.e., an approximated bound to $||\bm{a}_\text{W}||_2\leq1$ \SI{}{\meter/\second^2}. All evaluations have been conducted on an Intel Core i7-10850H notebook.

\subsection{Simulation results}
\noindent Regarding stage 1, the spline of PH nonic curves obtained from solving the OCP in \eqref{eq:ph_ocp} is depicted on the left side of Fig.~\ref{fig:results}. Its cost function value $f_\text{PH}$ in~\eqref{eq:ph_costfun} is $3.56\times10^{-5}$, indicating that along the entire curve $\rchi_1\approx0$, i.e., its adapted frame approximates an RMF. Intuitively, the moving frame portrayed in Fig.~\ref{fig:results} appears to be free of unwanted twists along the curve's tangent. In addition, the underlying quaternion polynomials, as well as their respective derivatives, are also displayed in the first four rows at the right side of Fig.~\ref{fig:results}. Notice that these are plotted against path-parameter $\xi$, and thus, each vertical line depicts the transition between two successive sections of the spline. Taking this into account, these four graphs illustrate the aforementioned requirement that the quaternion polynomial $\bm{Z}(\xi)$ has to be $C^3$ in order for the path-related functions $\sigma({\xi})$, $\text{R}(\xi)$, and $\bm{\omega}(\xi)$ to remain $C^2$ across the spline intersections. 

\begin{table}[t]
	\centering
	\caption{Average computation times in milliseconds required for solving each stage. Notice that the initialization is not necessary if a prior solution is available.} \label{tab:computation_times}
	\begin{tabular}{|M{22mm}|M{22mm}|M{22mm}|}
	    \hline
	    \multicolumn{2}{|c|}{Stage 1} &  Stage 2 \\
		\hline
		Initialization$^*$ & Solve~\eqref{eq:ph_ocp} & Solve~\eqref{eq:mpc_NLP} \\
		\hline
		\SI{27}{\milli\second} & \SI{115}{\milli\second}  & \SI{4}{\milli\second}\\
		\hline
	\end{tabular}
\end{table}

When it comes to stage 2, the trajectory computed from solving the NLP \eqref{eq:mpc_NLP} in a receding horizon fashion is shown by the colored line in the left column of Fig.~\ref{fig:results}. The associated color mapping refers to the norm of the acceleration. The corresponding system states and inputs are attached in the last three rows of the right column. Notice that these states do not reach the end of the path, because the simulation is stopped as soon as the last node gets to the end of the free space. Given the behavior incited by the progress maximization cost function, the system fully exploits the actuation by seeking the optimal trade-off between speed and spatial bounds. This phenomenon is observable in the last graph of Fig.~\ref{fig:results}, where at least one of the acceleration components is saturated throughout the majority of the curve, as well as in the top-view and side-view, where the planned trajectory remains within the free space while (nearly) touching the apex of the curves.

The respective computation times for both stages are listed in Table \ref{tab:computation_times}.  Given that the initialization is only necessary in the first iteration, i.e., when no prior solution exists, stage~1 can run in between $5$ to $10$ Hz, whereas stage~2 can be executed at up to $250$ Hz. These timings allow for the deployment of our approach in a variety of motion planning tasks. 

\section{CONCLUSION}\label{sec:conclusion}
\noindent In this work, we proposed a spatial motion planning control approach that allows to embed and leverage the environment's geometric properties in a prediction-based controller. For this purpose, we rely on a novel path-parameterization --agnostic to the path's adapted frame-- of the three dimensional Euclidean coordinates. For an efficient usage of this model, we exploit the properties of PH curves, which allow for representing the environment in terms of smooth functions that are dependent on a small number of parameters. Taking this into account, we suggest a hierarchical scheme for geometrically constrained motion planning, where the coefficients obtained from computing a collision-free spline of PH curves are fed into an NMPC scheme, whose plant model is based on the aforementioned path-parameterization. The presented scheme has been evaluated in an illustrative example within a highly non-convex free space and a progress-maximization cost function. Results suggest that our approach not only efficiently converts the geometric properties of the environment into an approximated rotation minimizing PH spline, but it also exploits it to approximate a desired performance criterion.
\vspace{1mm}
\section*{ACKNOWLEDGEMENTS}
\noindent The authors would like to thank Prof. Carolina Vittoria Beccari and Prof. Gudrun Albrecht, for the valuable help and discussions on spatial PH splines.
\vspace{1mm}

\newcommand{\BIBdecl}{\setlength{\itemsep}{0.45 em}} 
\bibliographystyle{IEEEtran}
\bibliography{spatial_motion_planning_with_pythagorean_hodograph_curves}

\end{document}